\documentclass[11pt]{article}

\usepackage[preprint]{acl}

\usepackage{times}
\usepackage{latexsym}
\usepackage{amsmath,amssymb}

\usepackage[T1]{fontenc}

\usepackage[utf8]{inputenc}

\usepackage{microtype}
\usepackage{inconsolata}

\usepackage{graphicx}
\usepackage{flafter}
\usepackage{stfloats}
\usepackage[below]{placeins}
\usepackage{float}
\title{Dynamic Adversarial Fine-Tuning Reorganizes Refusal Geometry}

\author{
  Wenhao Lan\textsuperscript{1} \quad
  Shan Li\textsuperscript{2} \quad
  Xinhua Lai\textsuperscript{1} \quad
  Meiqi Wu\textsuperscript{3} \quad
  Junbin Yang\textsuperscript{1} \quad
  Haihua Shen\textsuperscript{1} \quad
  Yijun Yang\textsuperscript{4} \\
  \textsuperscript{1}University of Chinese Academy of Sciences, Beijing, China \\
  \textsuperscript{2}Inner Mongolia University of Technology, Inner Mongolia, China \\
  \textsuperscript{3}Tsinghua University, Beijing, China \\
  \textsuperscript{4}Shandong University, Shandong, China \\
  {\normalfont\small\textbf{Correspondence:} Haihua Shen (\texttt{shenhh@ucas.ac.cn}) \enspace|\enspace Yijun Yang (\texttt{yijunyang@sdu.edu.cn})}
}

\begin{document}
\maketitle

\begin{abstract} 
Safety-aligned language models must refuse harmful requests without broad over-refusal, but it remains unclear how dynamic adversarial fine-tuning changes refusal-control carriers: Kullback--Leibler (KL)-constrained directions or small subspaces that causally modulate refusal without large safe-prompt distribution shifts. We study a 7B backbone under supervised fine-tuning (SFT) and Robust Refusal Dynamic Defense (R2D2), aligning HarmBench, StrongREJECT, and XSTest evaluations with five-anchor geometry measurements, causal interventions, and sparse adaptive stress tests. R2D2 drives fixed-source HarmBench attack success to zero at early checkpoints; however, these checkpoints also exhibit maximal XSTest refusal and fail a benign-utility audit. Later checkpoints partially recover utility-facing behavior while reopening attack success, with adaptive GCG attack success rate rising to 0.415 at step 250 and 0.613 at step 500. Internally, R2D2 preserves a late-layer admissible refusal-control carrier through step 100 and then relocates the best admissible carrier to an early layer; SFT relocates earlier yet remains less robust. Effective rank stays near 1.24, and SFT shows larger principal-angle drift, arguing against both dimensional expansion and drift magnitude as sufficient explanations. Causal interventions support a low-dimensional but utility-coupled carrier. These results support a geometry-reorganization account of R2D2 along a robustness--utility frontier, without establishing adaptive robustness. 
\end{abstract}

\section{Introduction}

A refusal policy is useful only if it is selective. A language model that refuses every prompt containing risky vocabulary may appear robust under a jailbreak metric, but it has not solved the safety problem: it has merely traded harmful compliance for broad non-compliance. This distinction matters during post-training, where standard supervised fine-tuning can weaken safety behavior, while adversarial fine-tuning can push the model toward refusal-heavy responses \citep{pham2025safeFinetune,mazeika2024harmbench}. Recent work on over-refusal and exaggerated safety has made the same point from the evaluation side: safe prompts that resemble unsafe ones should still be answered \citep{rottger2024xstest,cui2025orbench,pan2025overrefusal,brahman2024art}. Robust refusal therefore has to be read together with utility-facing behavior, not as an attack-success number in isolation.

This tension is especially sharp for dynamic adversarial fine-tuning. HarmBench introduced Robust Refusal Dynamic Defense (R2D2), which fine-tunes on a pool of adversarial cases that is repeatedly updated with GCG-style attacks \citep{zou2023universal,mazeika2024harmbench}. Unlike static supervised fine-tuning, this loop changes the model while also changing the pressure applied to it. A checkpoint can become harder to jailbreak because it has learned a better boundary, because it refuses more broadly, or because the internal mechanism controlling refusal has moved into a different part of the representation space. These possibilities look similar if we only evaluate outputs. They are not the same mechanism.

A separate line of work has shown that refusal is not only a surface behavior. Refusal can be modulated by directions in the residual stream \citep{arditi2024refusal}; jailbreaks and safety failures can be studied through hidden-state representations \citep{lin2024representationJailbreak,zhou2024alignmentJailbreak}; sparse and template-independent methods can recover refusal-related features or directions with causal effects \citep{yeo2025sparseRefusal,siu2025cosmic}. More recent geometry work further suggests that refusal may be better described by small subspaces or structured regions than by a single universal vector \citep{wollschlager2025geometry}. These results make refusal geometry measurable and sometimes intervenable. What they do not yet explain is how such geometry changes while a dynamic adversarial training loop is reshaping both the model and the attacks it sees.

The natural explanation would be simple drift: a refusal direction rotates during fine-tuning, and larger or smaller drift explains robustness. We find that this story is too weak for R2D2. In our matched supervised fine-tuning (SFT) trajectory, the relevant directions can drift more while the model remains substantially less robust. Conversely, R2D2 can maintain strong fixed-source robustness at early checkpoints without showing evidence that refusal has expanded into a higher-dimensional, redundant mechanism. This motivates a different unit of analysis. Rather than asking only where one refusal direction moves, we track the \emph{admissible refusal-control carrier}: Kullback--Leibler (KL) constrained refusal without large safe-prompt distribution shifts.

\begin{figure*}[t]
\centering
\includegraphics[width=\textwidth]{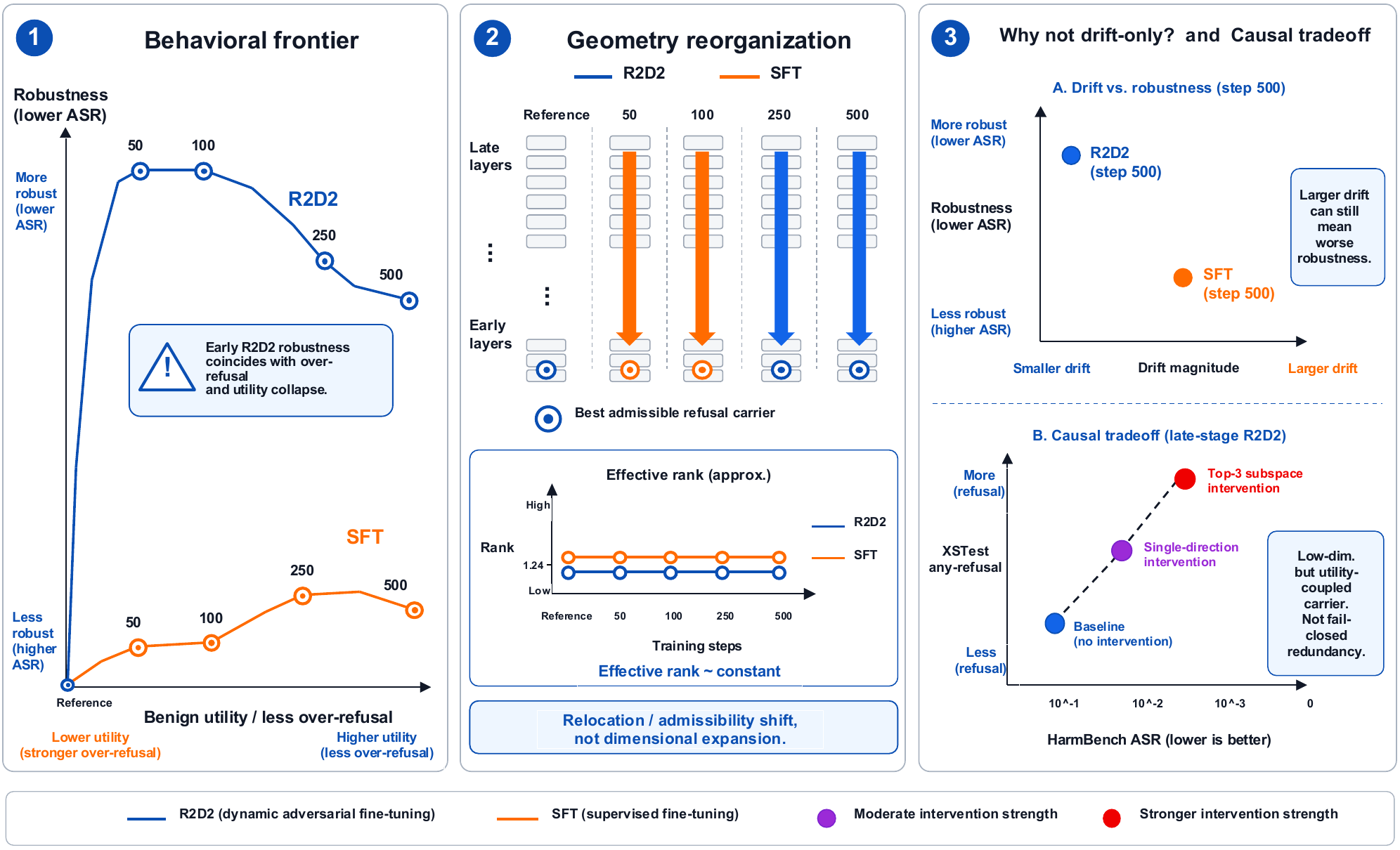}
\caption{Overview of the main finding. R2D2 moves along a robustness--utility frontier while its admissible refusal-control carrier reorganizes. Early checkpoints close fixed-source HarmBench attacks but over-refuse; later checkpoints recover some benign behavior while reopening attacks. The best carrier relocates across layers with near-constant effective rank, and causal interventions reveal low-dimensional but utility-coupled control.}
\label{fig:intro_frontier_teaser}
\end{figure*}

We study this process on a single Mistral-7B backbone under two fine-tuning regimes: standard supervised fine-tuning (SFT) and R2D2. The experimental design aligns behavior, geometry, and intervention at the same training anchors. On the behavior side, fixed-source HarmBench measures jailbreak attack success, StrongREJECT provides a complementary harmfulness-oriented evaluation, and XSTest measures refusal on safe prompts. We also run a sparse adaptive stress test on R2D2 checkpoints using checkpoint-specific GCG and AutoDAN attacks, and we add a small benign-utility audit to distinguish non-refusal from useful answering. On the geometry side, a five-anchor suite tracks admissible carriers, local rank, layer and position changes, and principal-angle drift. Finally, causal interventions test whether the selected directions or subspaces actually control refusal-side behavior.

The resulting trajectory is not monotone. Figure~\ref{fig:intro_frontier_teaser} summarizes the main pattern. R2D2 drives fixed-source HarmBench attack success to zero at early checkpoints, but those checkpoints also show maximal XSTest refusal and fail the benign-utility audit. Later checkpoints recover some utility-facing behavior, but this recovery coincides with reopened attack success. Sparse adaptive attacks make the late checkpoints look even less secure: the early checkpoint remains closed under the tested adaptive attacks, whereas later checkpoints reopen sharply. Internally, R2D2 preserves a late-layer admissible carrier through early training and then shifts the best admissible carrier to an early layer. The shift is not accompanied by dimensional expansion: effective rank remains close to constant. Nor is it explained by drift magnitude alone: SFT exhibits larger principal-angle drift while remaining less robust. Causal interventions complete the picture. A small carrier can restore refusal behavior at late R2D2 checkpoints, but the same intervention increases over-refusal and damages benign utility.

Our contributions are threefold.
\begin{enumerate}
    \item We provide a checkpoint-level measurement account of dynamic adversarial fine-tuning, connecting fixed-source robustness, sparse adaptive stress tests, over-refusal, benign utility, admissible-carrier geometry, and causal intervention in one trajectory.
    \item We show that R2D2 does not simply strengthen or rotate a fixed refusal direction. Instead, it reorganizes a low-dimensional refusal-control carrier while moving the model along a robustness--utility frontier.
    \item We provide evidence against two tempting explanations of robust refusal under R2D2: dimensional expansion and drift magnitude. The best admissible carrier relocates across training, effective rank remains nearly constant, and the carrier remains tightly coupled to benign utility.
\end{enumerate}

\section{Related Work}

\paragraph{Evaluating robust refusal and over-refusal.}
Recent robust-refusal evaluation separates attack success, harmful usefulness, and over-refusal, which earlier template-based jailbreak scores often conflated. HarmBench provides a standardized framework for automated red teaming and robust refusal, and introduces R2D2 as an adversarial fine-tuning procedure built around GCG-style updates \citep{zou2023universal,mazeika2024harmbench}. JailbreakBench makes a similar case for reproducible artifacts, explicit threat models, and standardized scoring \citep{chao2024jailbreakbench}. StrongREJECT targets a different weakness in jailbreak evaluation: a response should count as a successful jailbreak only when it gives useful harmful information, not merely when it avoids a refusal template \citep{souly2024strongreject}. Complementary work studies the other side of selective refusal: models may reject safe prompts that merely resemble unsafe ones. XSTest and OR-Bench measure over-refusal on benign prompts that safety-aligned models should answer \citep{rottger2024xstest,cui2025orbench}. Related work on safety decision boundaries and contextual noncompliance studies when models should comply, refuse, or explain noncompliance under more ambiguous contexts \citep{pan2025overrefusal,brahman2024art}. We use these benchmarks as behavioral readouts of a training trajectory: fixed-source HarmBench for attack success, StrongREJECT for harmful usefulness, and XSTest for exaggerated refusal. They show what the model does, but not which internal object changed.

\paragraph{Refusal directions in activation space.}
Mechanistic studies have made refusal an intervention target rather than only an output label. A central result is that adding or removing a residual-stream direction can induce or suppress refusal across several aligned models \citep{arditi2024refusal}. Other representation-space analyses connect jailbreak behavior to hidden-state separability and intermediate safety states \citep{lin2024representationJailbreak,zhou2024alignmentJailbreak}. More recent work extracts refusal latents with sparse autoencoders, identifies compliance directions from jailbreak initializations, and manipulates internal refusal stance \citep{yeo2025sparseRefusal,levi2025compliance,fu2025stance}. COSMIC is especially relevant for checkpoint studies because it selects refusal directions and layers using activation-space criteria rather than output refusal strings \citep{siu2025cosmic}. These methods give us tools for measuring refusal-related directions. Our setting asks a different question: how a dynamic adversarial training loop changes the admissible direction or small subspace that remains usable for controlling refusal.

\paragraph{From one direction to local refusal geometry.}
The single-direction view has also been refined. Concept-cone and representational-independence analyses argue that refusal can involve multiple directions or structured regions of activation space, and that orthogonality alone does not imply mechanistic independence \citep{wollschlager2025geometry}. This motivates measuring local rank, principal angles, admissibility, and intervention effects rather than treating one vector as the full mechanism. A related distinction is between harmfulness recognition and refusal behavior: recent work suggests that models can encode harmfulness separately from the decision to refuse \citep{zhao2025harmfulnessrefusal}. We therefore describe our object as a refusal-control carrier, not as the model's complete harmfulness representation. This scope is important for interpreting our geometry measurements and causal interventions.

\paragraph{Fine-tuning, adversarial training, and safety drift.}
Post-training can move both behavior and internal representations. Low-budget adaptation and layer-specific editing show that safety behavior can be weakened or modified by targeted updates \citep{qi2024fine,pham2025safeFinetune,zhao2024layerspecific}. DeepRefusal and related work treat refusal directions as trainable or ablatable components of the safety mechanism \citep{xie2025deeprefusal}. Projection-constraint work studies refusal-direction drift during tuning and attempts to stabilize it with an anchoring loss \citep{du2025anchoring}. ReFAT uses refusal-feature ablation as a training-time proxy for adversarial pressure \citep{yu2025refat}. R2D2 differs from these settings because the adversarial cases are refreshed during training, coupling parameter updates with a moving attack pressure \citep{mazeika2024harmbench}. This makes it a useful setting for studying geometry over time. Our contribution is not another drift metric or another defense; it is a checkpoint-level account of how R2D2 reorganizes refusal-control geometry, and how that reorganization aligns with robustness, over-refusal, utility, and causal controllability.

\section{Method and Experimental Setup}
\label{sec:methods}

Our goal is to measure how a common 7B base model evolves under standard supervised fine-tuning and dynamic adversarial fine-tuning. The behavioral comparison is cross-regime: each checkpoint is evaluated with the same external benchmarks and, for the main robustness axis, the same fixed-source attack set. The geometry comparison is more conservative. Because SFT and R2D2 use regime-native frozen candidate libraries, we interpret carrier trajectories within each regime and do not compare absolute geometry scores as if they came from a single shared calibration scale.

\subsection{Training regimes and anchors}

All experiments use Mistral-7B-v0.1 \citep{jiang2023mistral}. The SFT baseline is a LoRA-based supervised fine-tuning run on the UltraChat-200k instruction corpus. The R2D2 run follows the HarmBench dynamic adversarial training loop \citep{zou2023universal,mazeika2024harmbench}: a persistent pool of harmful test cases is maintained during training, a subset of cases is updated with GCG, and the model is optimized with clean SFT, away, and toward-refusal losses. Both runs use the same backbone, LoRA budget, training horizon, and seed; full optimizer, batching, adapter, checkpoint, and compute details are given in Appendix~\ref{app:method_details}.

Measurements are taken in two ways. First, a dense probe logs lightweight trajectory diagnostics throughout training, including the selected layer and position, direction norm, projection gap, cosine-to-anchor, and mini-COSMIC-style proxy statistics. These traces are used as monitors rather than as the main mechanistic evidence. Second, the main analysis uses a sparse five-anchor suite at the reference checkpoint and steps 50, 100, 250, and 500. The reference checkpoint for each regime is the earliest checkpoint at which the model exhibits stable direct-refusal behavior on the probe set: step 30 for R2D2 and step 5 for SFT\@. At each anchor we run the full admissible-carrier ranking, local low-rank analysis, principal-angle diagnostics, and the behavior evaluations described below. Causal interventions are run at steps 50, 250, and 500 for both regimes. This design keeps the training trajectory visible while reserving the strongest geometry and intervention tests for a small set of comparable checkpoints.

\subsection{Behavioral evaluations}

Fixed-source HarmBench GCG attack success rate (ASR) is the main robustness metric. ``Fixed-source'' means that attack cases are generated once from a designated source checkpoint using a fixed behavior set, attack configuration, random seed, and evaluator, and then replayed across checkpoints. This makes the checkpoint trajectory directly comparable, but it should not be read as adaptive worst-case robustness. StrongREJECT provides a complementary harmfulness-oriented score by asking whether outputs contain useful forbidden information rather than merely avoiding a refusal template \citep{souly2024strongreject}. XSTest measures over-refusal on safe prompts that resemble unsafe ones \citep{rottger2024xstest}.

Because XSTest non-refusal is not the same as useful assistance, we add a 60-prompt benign-utility sanity audit. Prompts cover everyday assistance, reasoning, writing, extraction, coding, and XSTest-safe cases. Each output is annotated on a three-point scale: 2 = clearly helpful, 1 = partially helpful, and 0 = refusal, empty or degenerate response, clearly wrong response, or off-task behavior. Outputs were manually annotated under the same fixed rubric by three annotators, with checkpoint labels and, for intervention outputs, intervention labels masked. Because this audit is intended as a sanity check rather than a broad utility benchmark, we report aggregate rates and do not treat it as a comprehensive capability evaluation.

The fixed-source protocol is the main trajectory estimand because it holds the attack distribution fixed while the checkpoint changes. To test how much of this picture survives attack regeneration, we also run a sparse adaptive stress test on R2D2 at steps 50, 250, and 500. For each selected checkpoint, GCG and AutoDAN attacks are regenerated over 400 HarmBench behaviors \citep{zou2023universal,liu2024autodan}. These runs are reported as a pressure test only: they are R2D2-only, sparse in training time, and limited to two attack families.

\subsection{Admissible refusal-control geometry}

We impose admissibility because a direction that separates harmful and safe activations can still be a poor control carrier if it causes large safe-prompt distributional shifts. For each regime, let $\mathcal{L}$ be the frozen, pruned candidate-layer set used by the anchor suite, and let $\mathcal{P}=\{-5,-4,-3,-2,-1\}$ be the post-instruction token-position grid. Following prior refusal-direction selection protocols~\citep{siu2025cosmic,arditi2024refusal}, we exclude candidate directions from the final 20\% of transformer blocks before selecting admissible carriers. Thus, all ``best carrier'' statements are conditional on this fixed pruned search space.

For checkpoint $t$, token position $p\in\mathcal{P}$, and layer $\ell\in\mathcal{L}$, we form the harmful-minus-safe residual-stream contrast
\begin{equation}
\label{eq:carrier}
r_{p,\ell}^{(t)}=
\mathbb{E}_{x\in\mathcal{D}_{\mathrm{harm}}}h_{p,\ell}^{(t)}(x)
-
\mathbb{E}_{x\in\mathcal{D}_{\mathrm{safe}}}h_{p,\ell}^{(t)}(x).
\end{equation}
The raw contrasts are used for spectral diagnostics; unit-normalized contrasts are used for ranking and intervention. We do not treat this contrast as a pure harmfulness vector or as a complete refusal mechanism. It may mix harmfulness-related, refusal-related, and policy-control signals, so we call it a candidate \emph{refusal-control carrier} and validate it through constrained interventions.

A carrier is admissible only if the safe-prompt KL divergence under ablation is finite and no larger than $\tau_{\mathrm{KL}}=0.10$. Among admissible candidates, we select the position--layer pair maximizing a COSMIC-style score, $S_{\mathrm{refuse}}+S_{\mathrm{comply}}-\beta_{\mathrm{KL}}\mathrm{KL}$, with $\beta_{\mathrm{KL}}=1.0$. Detailed formulas are provided in Appendix~\ref{app:geometry_formulas}. For the selected carrier we report its position, layer, total score, KL, refusal-side score, and compliance-side score. The same grid and scoring rule are used for all anchors within a regime.

To distinguish relocation from dimensional expansion, we also stack the raw candidate contrasts and report top-$k$ explained energy, $k_{90}$, $k_{95}$, effective rank, participation ratio, and top-3 principal angles to the regime-specific reference. These quantities are diagnostics of the local carrier library, not semantic counts of independent refusal mechanisms. Full scoring, KL, spectral, and principal-angle definitions are given in Appendix~\ref{app:geometry_formulas}.

\subsection{Causal interventions and reporting scope}

For both regimes, causal tests are run at steps 50, 250, and 500. For each checkpoint, we evaluate two diagnostic ablations during generation. The first removes the selected unit carrier from residual-stream activations at the selected layer. The second removes the top-3 local subspace formed from all candidate positions at that same layer. These interventions test whether the selected carrier or its local subspace has behavioral control; they are analysis operators, not proposed deployment-time defenses.

For each intervention we evaluate fixed-source GCG ASR, XSTest refusal, and benign utility when available, always against the no-intervention checkpoint baseline. The intervention rows are interpreted under the fixed-source protocol, since their purpose is to probe control through the measured carrier rather than to estimate adaptive worst-case robustness. Full operator definitions are given in Appendix~\ref{app:causal_operators}.

All safety-sensitive outputs are reported in aggregate. The paper does not print raw harmful prompts, optimized suffixes, transferable jailbreak strings, harmful completions, or sparse-adaptive attack artifacts. This reporting choice makes the experimental design auditable and the aggregate claims interpretable without unnecessarily releasing attack material.

\section{Results}
\label{sec:results}

We report five results. Under the fixed protocol, R2D2 follows a robustness--utility frontier; sparse adaptive attacks sharpen the late-stage fail-open pattern; the best admissible carrier relocates without clear dimensional expansion; principal-angle drift is not sufficient to explain robustness; and small causal carriers remain strongly utility-coupled.

\subsection{R2D2 Traces a Robustness--Utility Frontier}
\label{sec:results-frontier}

\begin{figure}[!t]
\centering
\includegraphics[width=\linewidth]{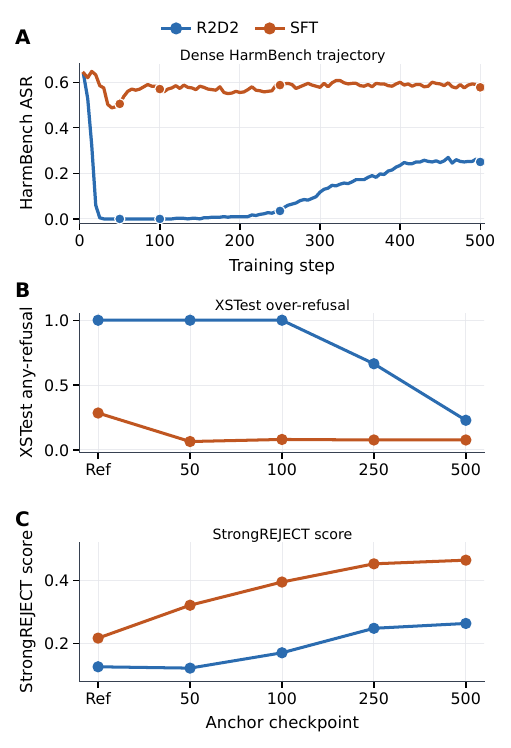}
\caption{\textbf{R2D2 traces a fixed-protocol robustness--over-refusal frontier.}
Panel A plots dense fixed-source HarmBench GCG ASR; Panel B plots XSTest any-refusal; Panel C plots StrongREJECT mean score. Early R2D2 closes fixed-source attacks but over-refuses; later checkpoints reduce over-refusal while partially reopening attack success.}
\label{fig:frontier_overview}
\end{figure}

R2D2 improves fixed-source robustness by first entering an extreme over-refusal regime, not by monotonically improving safety. Under fixed-source HarmBench GCG, R2D2 reaches ASR 0.000 at steps~50 and 100, then partially reopens to 0.035 at step~250 and 0.250 at step~500. SFT remains much less robust at the same anchors, with ASR 0.505, 0.570, 0.588, and 0.578. XSTest moves in the opposite direction: R2D2 any-refusal is 1.000 at the reference, step~50, and step~100, then drops to 0.664 at step~250 and 0.228 at step~500. StrongREJECT gives a consistent endpoint contrast: from the reference to step~500, the score changes from 0.126 to 0.264 for R2D2 and from 0.217 to 0.465 for SFT\@. Together, the anchor-level contrast and dense HarmBench/XSTest/StrongREJECT trajectories trace a frontier rather than monotonic safety improvement (Table~\ref{tab:behavior_summary}; Figure~\ref{fig:frontier_overview}).

Direct utility annotations show why XSTest non-refusal is not enough. On the 60-prompt benign sanity set, R2D2 reference and step~50 have strict utility, lenient utility, and mean helpfulness all equal to 0.000. By step~250, R2D2 reaches strict utility 0.150, lenient utility 0.633, and mean helpfulness 0.783; by step~500, it reaches 0.250, 0.867, and 1.117. Thus early R2D2 robustness is paired with zero scored utility on this benign sanity set, while late utility-facing recovery is partial and coincides with partial fail-open behavior. Appendix Table~\ref{tab:utility_baseline} gives the full annotation table, and Appendix Figure~\ref{fig:utility_trajectory} plots the same utility trajectory.

\begin{table}[t]
\centering
{\scriptsize
\setlength{\tabcolsep}{3pt}
\renewcommand{\arraystretch}{0.98}
\begin{tabular}{lcccc}
\hline
Anchor & \shortstack{R2D2\\HB ASR} & \shortstack{SFT\\HB ASR} & \shortstack{R2D2\\XS any-ref.} & \shortstack{SFT\\XS any-ref.} \\
\hline
Reference & 0.000 & 0.640 & 1.000 & 0.284 \\
50 & 0.000 & 0.505 & 1.000 & 0.064 \\
100 & 0.000 & 0.570 & 1.000 & 0.080 \\
250 & 0.035 & 0.588 & 0.664 & 0.076 \\
500 & 0.250 & 0.578 & 0.228 & 0.076 \\
Dense mean ASR & 0.114 & 0.579 & -- & -- \\
\hline
\end{tabular}
}
\caption{\textbf{Anchor-level behavior under the fixed protocol.}
Fixed-source HarmBench GCG ASR and XSTest any-refusal show R2D2 closing attacks early only by entering maximal over-refusal; later anchors reduce over-refusal while reopening ASR. ``Reference'' is the regime-specific aligned checkpoint used by the geometry suite. Dense mean ASR is computed over the dense trajectory.}
\label{tab:behavior_summary}
\end{table}

\subsection{Sparse Adaptive Attacks Sharpen the Frontier}
\label{sec:results-sparse-adaptive}

Sparse adaptive attacks show that the late utility-facing recovery is weaker than the fixed-source trajectory alone suggests. At step~50, regenerated GCG and AutoDAN attacks both have ASR 0.000, matching the fixed-source result; this checkpoint, however, is also the utility-collapsed regime in Table~\ref{tab:utility_baseline}, so the zero-ASR should not be read as a useful robust model.

At later checkpoints, regenerating attacks exposes a sharper fail-open pattern. At steps~250/500, fixed-source GCG ASR is 0.035/0.250, while adaptive GCG rises to 0.415/0.613 and AutoDAN rises to 0.158/0.270. Fixed-source evaluation remains the more controlled axis for aligning behavior with geometry across a dense trajectory, but it can underestimate late-stage vulnerability under checkpoint-specific attacks. Appendix Table~\ref{tab:sparse_adaptive} gives the complete sparse stress-test table.

\subsection{Geometry Changes Reflect Reorganization Rather than Dimensional Expansion}
\label{sec:results-geometry}

The geometry evidence is best described as carrier relocation and admissibility shift. R2D2's best KL-admissible carrier remains late-layer through step~100: \((\mathrm{pos}=-4, \mathrm{layer}=24)\) at the reference checkpoint, \((\mathrm{pos}=-4, \mathrm{layer}=23)\) at step~50, and \((\mathrm{pos}=-4, \mathrm{layer}=24)\) at step~100. By steps~250 and 500, the best KL-admissible carrier relocates to \((\mathrm{pos}=-3, \mathrm{layer}=0)\). SFT reaches the early-layer admissible regime earlier, already at step~50, while remaining less robust under fixed-source HarmBench. This statement is about the best KL-admissible intervention carrier, not the disappearance of all late-layer separation: late-layer candidates can remain discriminative while becoming less usable under the fixed admissibility constraint.

\begin{figure}[!t]
\centering
\includegraphics[width=\linewidth]{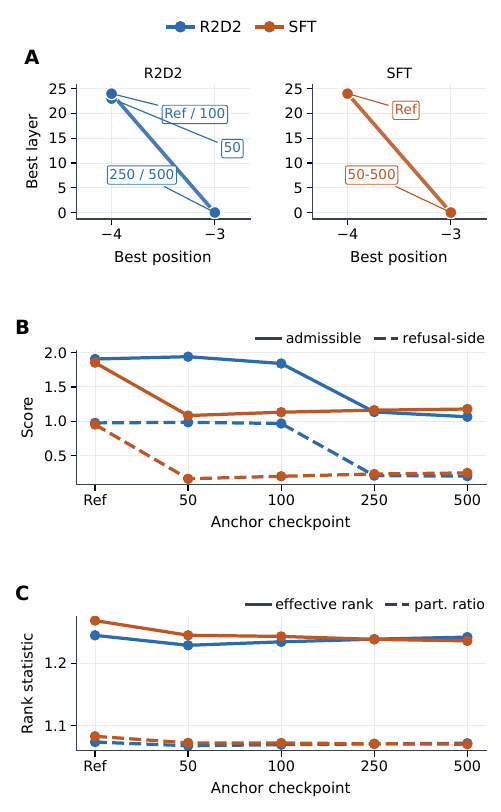}
\caption{\textbf{Refusal geometry changes by carrier relocation, not dimensional expansion.}
Panel A plots the best KL-admissible carrier; Panel B plots total and refusal-side scores; Panel C plots effective rank and participation ratio. R2D2 keeps a high-scoring late-layer carrier through step~100 and shifts the best admissible carrier to an early layer by step~250, while rank statistics remain nearly flat.}
\label{fig:geometry_reorganization}
\end{figure}

The relocation is not accompanied by clear dimensional growth. For R2D2, the best admissible score remains high at the reference, step~50, and step~100 anchors (1.91, 1.94, and 1.84) and falls at steps~250 and 500 (1.13 and 1.07); the refusal-side component drops from about 0.97 to about 0.20. Across both regimes, effective rank stays near 1.23--1.27, participation ratio changes little, and \(k_{90}\) and \(k_{95}\) remain 1. The supported claim is therefore reorganization under a low-rank local structure, not high-dimensional expansion. This relocation-without-expansion pattern is summarized in Figure~\ref{fig:geometry_reorganization}.

\subsection{Drift Magnitude Does Not Explain Robustness}
\label{sec:results-drift}
Principal-angle drift provides a useful negative control, but it does not explain robustness by itself. At step~500, the largest top-3 principal angle is \(42.55^\circ\) for R2D2 and \(53.31^\circ\) for SFT, yet fixed-source HarmBench ASR is lower for R2D2 (0.250) than for SFT (0.5775). Larger subspace rotation therefore does not imply stronger robust refusal. The relevant change is which carrier remains admissible, where that carrier is located, and how it behaves under intervention. Appendix Figure~\ref{fig:drift_vs_robustness} gives the full drift diagnostic.

\subsection{Causal Interventions Reveal Low-Dimensional but Utility-Coupled Carriers}
\label{sec:results-causal}

Late-stage R2D2 behavior is controlled by a small carrier, but stronger control pushes the model back toward broad refusal. At step~250, top-3 subspace ablation reduces fixed-source ASR from 0.035 to 0.0000 and raises XSTest any-refusal from 0.664 to 0.988. At step~500, it reduces ASR from 0.250 to 0.0000 and raises XSTest any-refusal from 0.228 to 0.968. Single-direction ablation is also strong, reducing ASR to 0.0031 at both late anchors. Appendix Table~\ref{tab:causal_behavior} gives the full intervention table.

\begin{figure}[!tbp]
\centering
\includegraphics[width=\linewidth]{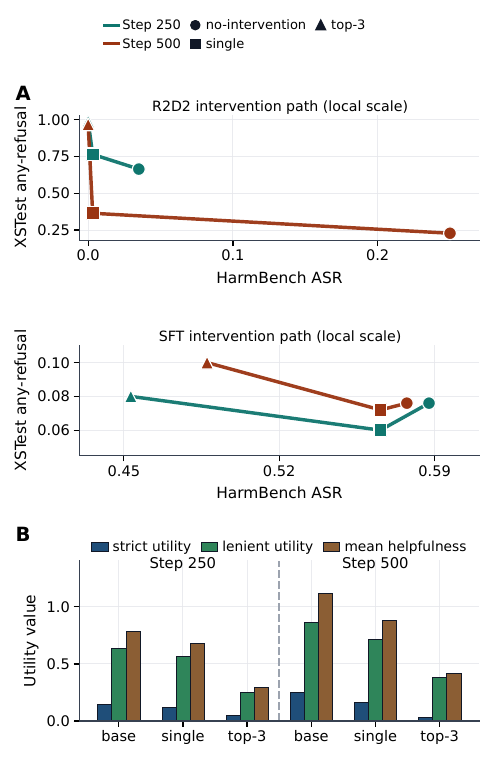}
\caption{\textbf{Late-stage R2D2 carriers are low-dimensional but utility-coupled.}
Panel A compares no intervention, single-direction removal, and top-3 subspace removal for R2D2 and SFT at steps 250/500 in HarmBench ASR versus XSTest refusal space. Panel B gives matched R2D2 benign-utility annotations. Small subspace interventions nearly eliminate fixed-source ASR, but increase over-refusal and reduce benign utility.}
\label{fig:causal_tradeoff}
\end{figure}

The same interventions reduce benign utility. At step~250, top-3 ablation lowers R2D2 strict utility, lenient utility, and mean helpfulness from 0.150, 0.633, and 0.783 to 0.050, 0.250, and 0.300. At step~500, it lowers the same metrics from 0.250, 0.867, and 1.117 to 0.033, 0.383, and 0.417. SFT interventions are weaker on fixed-source ASR and do not produce the same utility collapse: at step~500, SFT top-3 ablation lowers ASR only from 0.5775 to 0.4875. These results support low-dimensional but utility-coupled control, not a deployable defense or fail-closed redundancy. The resulting ASR--over-refusal--utility tradeoff is summarized in Figure~\ref{fig:causal_tradeoff}.

At step~50, top-3 ablation gives a useful negative sanity check. It keeps R2D2 fixed-source ASR at 0.000 and drops XSTest any-refusal from 1.000 to 0.000, but strict utility, lenient utility, and mean helpfulness remain 0.000 while the empty-or-nonsensical rate reaches 1.000. This is degenerate non-refusal, not utility restoration. Appendix Table~\ref{tab:step50_top3_sanity} reports the row-level evidence.

\section{Discussion and Conclusion}
\label{sec:discussion}
\label{sec:conclusion}

Our results recast robust refusal as a behavior--geometry trajectory, not an endpoint ASR or a drifting vector. This complements single-direction and COSMIC-style accounts \citep{arditi2024refusal,siu2025cosmic}: they show that directions can modulate refusal; we ask which KL-admissible carrier remains usable while dynamic adversarial fine-tuning changes the model. The object is narrower than ``the refusal mechanism'' and more operational than a raw activation contrast.

The central pattern is reorganization. SFT drifts more in principal angle yet remains less robust, so drift magnitude is insufficient. R2D2 keeps a high-scoring late-layer admissible carrier during the early robust-but-over-refusing phase, then shifts the best carrier to an early layer as benign behavior partly returns and attack success reopens. Flat effective-rank and participation-ratio diagnostics point to relocation within a low-rank local structure, not dimensional expansion or independent fail-closed pathways.

This yields a robustness--utility frontier. Early R2D2 closes fixed-source attacks but over-refuses and fails the benign-utility audit; later checkpoints answer more benign prompts yet fail open, especially under regenerated attacks. Causal ablations make the cost visible: late-stage single-direction and top-3 interventions restore fixed-source robustness by increasing XSTest refusal and lowering benign utility. The step-50 sanity check separates non-refusal from usefulness because suppressing refusal-like outputs does not restore helpful behavior.

The claim is therefore bounded. Harmful--safe contrasts identify behaviorally active refusal-control carriers, not a full harmfulness representation, and the evidence does not establish adaptive robustness or fail-closed redundancy. Under this fixed protocol, dynamic adversarial fine-tuning reorganizes a low-dimensional, utility-coupled admissible carrier; robust refusal should be evaluated as a behavior--geometry trajectory, not by attack success or refusal strength alone.

\section*{Limitations}
\label{sec:limitations}

The evidence in this paper is intentionally narrow. We study one 7B backbone and two LoRA-based fine-tuning regimes, SFT and R2D2-style dynamic adversarial fine-tuning. Holding the model fixed makes the trajectory comparison easier to interpret, but the observed frontier may depend on the backbone, adapter configuration, training budget, checkpoint cadence, and seed. We therefore do not claim that other model families or alignment methods will exhibit the same geometric trajectory.

The geometry results should be read as protocol-level measurements. SFT and R2D2 are analyzed with regime-native frozen admissible libraries, so absolute geometry scores are not measurements from a single shared instrument. Our cross-regime comparisons rely on shared behavior evaluations and on within-regime geometry trajectories. Likewise, the best admissible carrier is a KL-constrained measurement object selected from a fixed candidate library; it is not a complete map of the refusal mechanism.

We do not separate harmfulness representations from refusal-control representations. The harmful--safe activation contrasts in Eq.~\eqref{eq:carrier} are used to select carriers that are validated through refusal-side behavioral interventions, but recent work suggests that harmfulness and refusal can be encoded separately \citep{zhao2025harmfulnessrefusal}. Our claims are therefore limited to admissible refusal-control carriers under this protocol, not to the full internal representation of harmfulness or safety belief.

The primary robustness trajectory uses fixed-source HarmBench GCG\@. This design fixes the attack distribution while checkpoints change, which is useful for trajectory analysis, but it is not an adaptive worst-case robustness estimate. We include a sparse adaptive stress test with GCG and AutoDAN at three R2D2 checkpoints, but it is still narrow: it is R2D2-only, sparse over training time, and limited to two attack families. StrongREJECT and XSTest add complementary harmfulness and over-refusal views, but they do not replace broad utility testing, SFT-side adaptive evaluation, black-box optimization, human jailbreak families, or multi-turn attack settings.

The intervention and utility analyses are diagnostic rather than exhaustive. Causal interventions are tested at selected anchors with single-direction and top-3 subspace ablations, so they do not rule out other layers, operators, prompts, or decoding settings. The 60-prompt benign-utility audit is useful for separating XSTest non-refusal from useful assistance, especially in the step-50 top-3 case, but it is not a general capability benchmark.

Finally, the results do not establish independent fail-closed refusal pathways or adaptive robustness. The current evidence supports low-dimensional but utility-coupled causal control under this fixed protocol, and the sparse adaptive runs show substantial late-stage vulnerability rather than a deployed-safe endpoint. Demonstrating fail-closed redundancy would require showing that harmful-request refusal persists when one carrier is removed while benign utility is preserved.

\section*{Ethical Considerations and Responsible NLP}
\label{sec:ethics}

This work studies jailbreak robustness and activation-space refusal interventions for defensive evaluation and measurement. The same analyses could be misused to identify weak checkpoints, attack-sensitive carriers, or intervention strategies that alter refusal behavior. We therefore report aggregate behavioral, geometric, causal-intervention, and sparse adaptive-attack metrics, but do not publish raw harmful prompts, optimized adversarial suffixes, harmful completions, transferable exploit strings, or unredacted attack logs.

Our release plan prioritizes auditability without lowering the cost of misuse. We will release sanitized plotting code, geometry-analysis code, figure-source tables, hyperparameter and compute summaries, benchmark-version information, and sanitized configuration files. The reported aggregate claims can be audited from the released tables, code, and configuration metadata; full retraining or re-evaluation would require rerunning the training and evaluation pipeline. We do not plan to publicly release trained adapters or checkpoints with attack-sensitive behavior; any future release of such artifacts will use access controls and usage terms appropriate to their misuse risk. Causal-intervention code is treated as analysis infrastructure rather than as a deployable defense or attack pipeline.

The study uses established safety benchmarks and a small benign-utility sanity set. Harmful-content artifacts are handled as evaluation data and are not reproduced in the paper or appendix. The benign-utility audit is used only to distinguish non-refusal from useful behavior; it does not require generating harmful content. Dataset licenses, benchmark access restrictions, evaluator versions, annotation rubrics, and withheld artifacts are documented in the checklist or artifact notes.

\bibliography{custom}
\clearpage
\appendix

\section{Training, Compute, and Measurement Details}
\label{app:method_details}

Table~\ref{tab:training_settings} reports the full training settings used for the two regimes. We place these details in the appendix so that the main method section can focus on the estimand, metrics, and intervention operators.

\begin{table*}[!b]
\centering
\scriptsize
\setlength{\tabcolsep}{3pt}
\begin{tabular}{p{0.24\textwidth}p{0.34\textwidth}p{0.34\textwidth}}
\hline
Setting & SFT & R2D2 \\
\hline
Backbone & Mistral-7B-v0.1 & Mistral-7B-v0.1 \\
Adapter & LoRA $r=64$, $\alpha=16$, dropout 0.1 & same \\
Target modules & q/k/v/o projections & same \\
Supervised data & UltraChat-200k train/test splits & same supervised component \\
Adversarial data & none & persistent HarmBench-derived test-case pool \\
Learning rate & $2\times10^{-5}$ & $2\times10^{-5}$ \\
Scheduler / optimizer & cosine / \texttt{adamw\_torch} & cosine / \texttt{adamw\_torch} \\
Warmup & \texttt{warmup\_ratio}=0; \texttt{warmup\_steps}=0 & same \\
Batching & per-device batch 1; gradient accumulation 16 & per-device batch 1; gradient accumulation 16; plus away/toward examples \\
Precision & bf16 & bf16 \\
Max length / horizon & 2048 tokens / 500 steps & 2048 tokens / 500 steps \\
Seed & 42 & 42 \\
Save cadence & every 5 steps & every 5 optimizer steps through the custom \texttt{save\_every} path \\
R2D2 GCG settings & n/a & $N=180$, $n=8$, $m=5$, width 512, reset 20\% every 50 steps \\
\hline
\end{tabular}
\caption{Training settings for the SFT and R2D2 runs. For R2D2, $N$ is the persistent test-case pool size, $n$ is the number of cases updated per optimizer step, $m$ is the number of GCG update steps per selected case, and ``width'' is the GCG search width.}
\label{tab:training_settings}
\end{table*}

The R2D2 implementation combines clean SFT with adversarial away and toward-refusal losses. In this run, the implemented objective is
\begin{equation}
\label{eq:r2d2-loss}
\mathcal{L}_{\mathrm{R2D2}}
=
\mathcal{L}_{\mathrm{clean}}
+0.5\,\mathcal{L}_{\mathrm{away}}
+0.5\,\mathcal{L}_{\mathrm{toward}}.
\end{equation}
For adversarial prompt $x_i^{\mathrm{adv}}$, harmful target tokens $y_{i,\tau}^{\mathrm{harm}}$, and fixed refusal-target tokens $y_{\tau}^{\mathrm{ref}}$, define
\begin{equation}
\label{eq:r2d2-token-probs}
\begin{aligned}
p_{i,\tau}^{h}
&=P_\theta(y_{i,\tau}^{\mathrm{harm}}\mid x_i^{\mathrm{adv}},y_{i,<\tau}^{\mathrm{harm}}),\\
p_{i,\tau}^{r}
&=P_\theta(y_{\tau}^{\mathrm{ref}}\mid x_i^{\mathrm{adv}},y_{<\tau}^{\mathrm{ref}}).
\end{aligned}
\end{equation}
The away term is the implementation's token-level thresholded \texttt{log\_1\_minus\_p} loss, while the toward term is token-level cross-entropy:
\begin{equation}
\label{eq:r2d2-away-toward}
\begin{aligned}
\mathcal{L}_{\mathrm{away}}
&=\frac{1}{Z_h}
\sum_{i,\tau} a_{i,\tau}
\,\ell_{\mathrm{l1mp}}(\log p_{i,\tau}^{h}),\\
\ell_{\mathrm{l1mp}}(s)
&=\mathbf{1}[s\ge -5]\big[-\log(1-\exp(s))\big],\\
\mathcal{L}_{\mathrm{toward}}
&=-\frac{1}{Z_r}
\sum_{i,\tau} b_{i,\tau}\log p_{i,\tau}^{r}.
\end{aligned}
\end{equation}
Here $a_{i,\tau}$ and $b_{i,\tau}$ are the non-padding token masks, and $Z_h$ and $Z_r$ are the corresponding token normalizers.

\subsection{Admissible-Carrier Scoring and Geometry Diagnostics}
\label{app:geometry_formulas}

This subsection gives the full formulas summarized in Section~\ref{sec:methods}. We use the raw contrasts in Eq.~\eqref{eq:carrier} for spectral analysis and the unit-normalized carrier
\begin{equation}
\label{eq:unit-carrier}
u_{p,\ell}^{(t)}=
\frac{r_{p,\ell}^{(t)}}{\|r_{p,\ell}^{(t)}\|_2+\epsilon_{\mathrm{num}}},
\qquad \epsilon_{\mathrm{num}}=10^{-12}
\end{equation}
for ranking and intervention. Let $A_t(x;\mathcal{I})$ be the normalized COSMIC-style evaluation activation for prompt $x$ after intervention $\mathcal{I}$, averaged over the evaluation-layer set. Let $c_{\mathrm{harm}}^{(t)}(z)=\cos(z,\mu_{\mathrm{harm}}^{(t)})$ and $c_{\mathrm{safe}}^{(t)}(z)=\cos(z,\mu_{\mathrm{safe}}^{(t)})$, where the centroids are computed from harmful/refusal-side and safe/compliance-side validation activations. For readability, $u$ denotes $u_{p,\ell}^{(t)}$, while $\mathcal{I}_{+}(u)$ and $\mathcal{I}_{-}(u)$ denote addition and ablation. The two ranking scores are
\begin{equation}
\label{eq:refuse-comply-scores}
\begin{aligned}
S_{\mathrm{refuse}}^{(t)}(p,\ell)
&=\mathbb{E}_{x\in\mathcal{D}_{\mathrm{safe}}^{\mathrm{val}}}
 c_{\mathrm{harm}}^{(t)}\!\left(A_t(x;\mathcal{I}_{+}(u))\right),\\
S_{\mathrm{comply}}^{(t)}(p,\ell)
&=\mathbb{E}_{x\in\mathcal{D}_{\mathrm{harm}}^{\mathrm{val}}}
 c_{\mathrm{safe}}^{(t)}\!\left(A_t(x;\mathcal{I}_{-}(u))\right).
\end{aligned}
\end{equation}
The first term tests whether adding the carrier moves safe prompts toward refusal-side activations; the second tests whether removing it moves harmful prompts toward compliance-side activations. These scores are used only to rank candidate carriers.

Admissibility uses the ablation-side safe-prompt KL\@. Writing $P_t^{\mathcal{I}}$ for the distribution under intervention $\mathcal{I}$ and $P_t$ for the no-intervention distribution,
\begin{equation}
\label{eq:safe-kl}
\begin{aligned}
\mathrm{KL}^{(t)}(p,\ell)
&=\frac{1}{|\mathcal{D}_{\mathrm{safe}}^{\mathrm{val}}|}
\sum_{x\in\mathcal{D}_{\mathrm{safe}}^{\mathrm{val}}}\\
&\quad D_{\mathrm{KL}}\!\left(
P_t(\cdot\mid x)\;\|\;
P_t^{\mathcal{I}_{-}(u)}(\cdot\mid x)
\right).
\end{aligned}
\end{equation}
A candidate is ablation-admissible when this side effect is finite and at most $\tau_{\mathrm{KL}}=0.10$:
\begin{equation}
\label{eq:admissible-set}
\begin{aligned}
\mathcal{A}^{(t)}=\{(p,\ell)\in\mathcal{P}\times\mathcal{L}:{}&
\mathrm{KL}^{(t)}(p,\ell)<\infty,\\
&\mathrm{KL}^{(t)}(p,\ell)\le\tau_{\mathrm{KL}}\}.
\end{aligned}
\end{equation}
Among ablation-admissible candidates, we select
\begin{equation}
\label{eq:carrier-selection}
(p_t^\star,\ell_t^\star)=
\arg\max_{(p,\ell)\in\mathcal{A}^{(t)}}
S_{\mathrm{total}}^{(t)}(p,\ell),
\end{equation}
where
\begin{equation}
\label{eq:total-score}
\begin{aligned}
S_{\mathrm{total}}^{(t)}(p,\ell)
&=S_{\mathrm{refuse}}^{(t)}(p,\ell)+S_{\mathrm{comply}}^{(t)}(p,\ell)\\
&\quad-\beta_{\mathrm{KL}}\mathrm{KL}^{(t)}(p,\ell),
\end{aligned}
\end{equation}
with $\beta_{\mathrm{KL}}=1.0$.

For spectral diagnostics, we stack the raw contrasts before unit normalization,
\begin{equation}
\label{eq:carrier-matrix}
R^{(t)}=
\begin{bmatrix}
(r_{p_1,\ell_1}^{(t)})^\top\\
\cdots\\
(r_{p_m,\ell_m}^{(t)})^\top
\end{bmatrix}
\in\mathbb{R}^{m\times d},
\quad m=|\mathcal{P}||\mathcal{L}|.
\end{equation}
Let $\{\sigma_i\}$ be its singular values and $q_i=\sigma_i^2/\sum_j\sigma_j^2$. We report top-$k$ explained energy, $k_{90}$, $k_{95}$, effective rank, and participation ratio, with
\begin{equation}
\label{eq:k-rho}
\begin{aligned}
k_\rho&=\min\left\{k:\sum_{i=1}^{k}q_i\ge\rho\right\},
\rho\in\{0.90,0.95\},
\end{aligned}
\end{equation}
and
\begin{equation}
\label{eq:spectral-diagnostics}
\begin{aligned}
\mathrm{erank}(R^{(t)})
&=\exp\!\left(-\sum_{i:q_i>0}q_i\log q_i\right),\\
\mathrm{PR}(R^{(t)})
&=\frac{1}{\sum_i q_i^2}.
\end{aligned}
\end{equation}
Finally, let $Q_k^{(t)}$ and $Q_k^{(\mathrm{ref})}$ contain the top-$k$ right singular vectors of $R^{(t)}$ and the regime-specific reference matrix. With $C_k^{(t)}=(Q_k^{(\mathrm{ref})})^\top Q_k^{(t)}$, the principal angles are
\begin{equation}
\label{eq:principal-angles}
\theta_i^{(t)}=\cos^{-1}\!\left(\sigma_i(C_k^{(t)})\right),
\qquad i=1,\ldots,k.
\end{equation}
We use $k=3$ for the reported drift diagnostic and report the angles in degrees.

\subsection{Causal-Intervention Operators}
\label{app:causal_operators}

For checkpoint $t$, let $u^\star=u_{p_t^\star,\ell_t^\star}^{(t)}$ be the selected unit carrier. During generation, the single-direction diagnostic removes this component from each residual-stream vector at the selected layer,
\begin{equation}
\label{eq:single-direction-ablation}
h_{\tau,\ell_t^\star}
\leftarrow
h_{\tau,\ell_t^\star}
-
\left((h_{\tau,\ell_t^\star})^\top u^\star\right)u^\star,
\qquad \forall\tau.
\end{equation}
For the top-3 diagnostic, we form a selected-layer local matrix using all candidate positions at the selected layer,
\begin{equation}
\label{eq:selected-layer-matrix}
R_{\ell_t^\star}^{(t)}=
\begin{bmatrix}
(r_{p_1,\ell_t^\star}^{(t)})^\top\\
\cdots\\
(r_{p_{|\mathcal{P}|},\ell_t^\star}^{(t)})^\top
\end{bmatrix},
\qquad p_i\in\mathcal{P}.
\end{equation}
Let $U^\star\in\mathbb{R}^{d\times3}$ contain the top three orthonormal right singular vectors of $R_{\ell_t^\star}^{(t)}$. The subspace intervention applies
\begin{equation}
\label{eq:top3-ablation}
h_{\tau,\ell_t^\star}
\leftarrow
h_{\tau,\ell_t^\star}
-
U^\star(U^\star)^\top h_{\tau,\ell_t^\star},
\qquad \forall\tau.
\end{equation}
These operators follow the activation-intervention tradition of testing behavioral control through learned directions \citep{li2023iti,arditi2024refusal}.

\subsection{Compute metadata}
\label{app:compute}

We report a canonical artifact-producing budget because it maps most directly to the final on-disk artifacts used in this manuscript. Table~\ref{tab:compute_metadata} breaks the canonical ledger and additional recovered spend down by result block. Additional recovered spend includes failed, cancelled, superseded, and debugging jobs associated with the same fixed-protocol evidence chain. The sparse adaptive stress test is reconciled at the shard and behavior level in Table~\ref{tab:sparse_adaptive}; its job-level GPU-hour accounting was not recovered in the same Slurm ledger and is therefore excluded from the compute totals below.

\begin{table*}[!htbp]
  \centering
  \scriptsize
  \setlength{\tabcolsep}{3pt}
  \begin{tabular}{p{1.8cm}p{1.8cm}p{1.6cm}p{1.6cm}p{6.2cm}}
    \hline
    Block & Canonical hardware & Canonical GPU-hours & Additional recovered spend & Reviewer note \\
    \hline
    SFT training & 1$\times$A800 & 2.176 & 1.464 & Stable canonical run. \\
    R2D2 training & 4$\times$H100 & 186.893 & 96.551 & Slurm \texttt{FAILED} only after the final checkpoint was saved; the failure occurred in a later probe-stage wrapper step. \\
    Fixed-source HarmBench & 2$\times$A800 & 48.944 & 296.406 & Prompt-library completion plus checkpoint sweeps. \\
    Anchor + causal/XSTest & Mostly 1$\times$A800 & 3.706 & 2.055 & Includes superseded completed jobs and lightweight paper-stage retries. \\
    StrongREJECT & 1$\times$A800 / 2$\times$H100 & 3.031 & 0.000 & Canonical runs are reported separately; no additional spend is broken out here because the small paper-stage retry ledger is already included above. \\
    Benign utility & 2$\times$A800 / 1$\times$A800 & 0.735 & 0.000 & The April~17,~2026 annotation and summary refresh leaves a minor run-level provenance gap after the April~16,~2026 GPU jobs. \\
    Sparse adaptive stress test & not recovered & n/a & n/a & 120 completed shards across GCG/AutoDAN at R2D2 steps~50/250/500; excluded from GPU-hour totals because the matching Slurm job accounting was not recovered. \\
    Total & --- & 245.486 & 396.476 & Canonical artifact-producing ledger plus additional recovered spend equals 641.962 GPU-hours across 69 recovered Slurm records; sparse adaptive shard accounting is reported separately. \\
    \hline
  \end{tabular}
  \caption{Compute metadata for the fixed-protocol evidence chain. Canonical GPU-hours count the best-supported artifact-producing jobs that map to the final on-disk outputs used in this manuscript. Additional recovered GPU-hours count failed, cancelled, superseded, and debugging jobs associated with the same evidence chain and are reported only for transparency. The sparse adaptive stress test is reconciled separately at the shard level and is excluded from the GPU-hour totals because matching Slurm accounting was not recovered.}
  \label{tab:compute_metadata}
\end{table*}

\subsection{Additional measurement details}
\label{app:measurement_details}

Dense monitoring uses a lightweight COSMIC-style probe over fixed harmful and safe prompt sets and logs trajectory diagnostics at high cadence. The five-anchor suite uses the same anchor set as the behavior evaluations and reports the best admissible carrier, KL value, refusal-side score, comply-side score, local SVD statistics, effective rank, participation ratio, and principal angles relative to the regime-specific reference checkpoint. The benign-utility sanity set contains 60 safe prompts from everyday assistance, reasoning, writing, extraction, coding, and XSTest-safe buckets. The causal intervention rows use the same decoding and evaluation settings as the corresponding no-intervention checkpoints, and all deltas are interpreted under the fixed-source protocol. The sparse adaptive rows use the same HarmBench behavior set but regenerate attacks for the selected R2D2 checkpoints; they are analyzed as a separate stress test rather than as part of the dense geometry-alignment axis.

For a checkpoint $\theta_t$ and a fixed attack set $\mathcal{B}$, HarmBench attack success rate is aggregated as
\begin{equation}
\label{eq:asr}
\mathrm{ASR}^{(t)}=
\frac{1}{|\mathcal{B}|}
\sum_{b\in\mathcal{B}}
\mathbf{1}\!
\left[J_{\mathrm{HB}}(M_{\theta_t}(b))=\mathrm{success}\right].
\end{equation}
For XSTest, we report any-refusal on the safe prompt set $\mathcal{X}$:
\begin{equation}
\label{eq:xstest-any}
\begin{aligned}
\mathrm{AnyRef}^{(t)}
&=\frac{1}{|\mathcal{X}|}
\sum_{x\in\mathcal{X}}
\mathbf{1}\!\left[
J_{\mathrm{full}}(M_{\theta_t}(x))\right.\\
&\qquad\left.\lor
J_{\mathrm{partial}}(M_{\theta_t}(x))
\right].
\end{aligned}
\end{equation}
For checkpoint or intervention condition $t$, given $n_t$ benign-utility annotations $a_i^{(t)}\in\{0,1,2\}$, we compute
\begin{equation}
\label{eq:utility}
\begin{aligned}
U_{\mathrm{strict}}^{(t)}
&=\frac{1}{n_t}\sum_{i=1}^{n_t}\mathbf{1}[a_i^{(t)}=2],\\
U_{\mathrm{lenient}}^{(t)}
&=\frac{1}{n_t}\sum_{i=1}^{n_t}\mathbf{1}[a_i^{(t)}\ge1],\\
\bar{U}^{(t)}
&=\frac{1}{n_t}\sum_{i=1}^{n_t} a_i^{(t)}.
\end{aligned}
\end{equation}
Refusal and empty-or-nonsensical rates are computed with the same averaging convention over the 60 annotated prompts for each condition.

\FloatBarrier
\setlength{\textfloatsep}{7pt plus 1pt minus 2pt}
\setlength{\floatsep}{5pt plus 1pt minus 1pt}
\setlength{\intextsep}{6pt plus 1pt minus 1pt}
\setlength{\abovecaptionskip}{3pt plus 1pt minus 1pt}
\setlength{\belowcaptionskip}{0pt plus 1pt minus 0pt}
\makeatletter
\setlength{\@fptop}{0pt}
\setlength{\@fpsep}{6pt plus 1fil}
\setlength{\@fpbot}{0pt plus 1fil}
\makeatother
\suppressfloats[t]
\section{Additional Robustness, Utility, Drift, and Intervention Diagnostics}
\label{app:additional_results}

These appendix results provide four diagnostic checks that support, but do not expand, the main claims. The sparse adaptive stress test checks whether the fixed-source trajectory understates vulnerability when attacks are regenerated at selected checkpoints. The benign-utility audit asks whether non-refusal corresponds to useful assistance. The principal-angle diagnostic asks whether subspace rotation alone explains robustness. The intervention tables ask whether causal refusal control is utility-preserving. Together, these diagnostics support the robustness--utility frontier and low-dimensional utility-coupled control, rather than a separate adaptive-robustness or fail-closed-redundancy claim.

\subsection{Sparse Adaptive Stress Test}
\label{app:sparse_adaptive}

Table~\ref{tab:sparse_adaptive} reports the complete sparse adaptive stress-test rows summarized in Section~\ref{sec:results-sparse-adaptive}. Each adaptive attack/checkpoint pair evaluates 400 HarmBench behaviors across 20 completed shards. The table is a scope check rather than a full adaptive robustness suite.

\begin{table}[!htbp]
\centering
\small
\setlength{\tabcolsep}{4pt}
\begin{tabular}{lccc}
\hline
Evaluation & Step 50 & Step 250 & Step 500 \\
\hline
Fixed-source GCG & 0.000 & 0.035 & 0.250 \\
Sparse adaptive AutoDAN & 0.000 & 0.158 & 0.270 \\
Sparse adaptive GCG & 0.000 & 0.415 & 0.613 \\
\hline
\end{tabular}
\caption{Sparse adaptive stress tests on selected R2D2 checkpoints. Values are ASR over 400 HarmBench behaviors. Step~50 remains at zero ASR under both regenerated attacks, but this is the utility-collapsed early regime. Steps~250 and~500 reopen substantially, especially under regenerated GCG\@. These rows stress-test the fixed-source conclusion; they are not a full adaptive-robustness evaluation.}
\label{tab:sparse_adaptive}
\end{table}

\subsection{Benign-Utility Audit}
\label{app:benign_utility}

\begin{figure}[!htbp]
\centering
\includegraphics[width=0.94\linewidth,trim=6pt 6pt 6pt 6pt,clip]{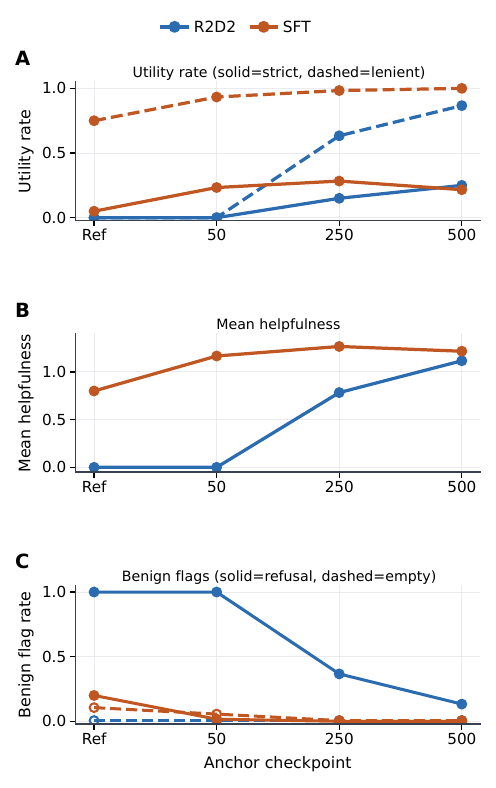}
\caption{Direct benign-utility trajectory on the 60-prompt sanity set. Panel A plots strict and lenient utility, using solid and dashed lines respectively. Panel B plots mean helpfulness. Panel C plots benign refusal and empty-or-nonsensical rates, again using solid and dashed lines respectively. The audit shows that early R2D2's closed fixed-source behavior coincides with zero benign utility.}
\label{fig:utility_trajectory}
\end{figure}

The benign-utility audit separates refusal, non-refusal, and useful assistance. Table~\ref{tab:utility_baseline} reports the raw annotations on the 60-prompt sanity set, and Figure~\ref{fig:utility_trajectory} plots the same trajectory. The early R2D2 checkpoints have zero strict utility, zero lenient utility, and zero mean helpfulness, which is why the closed fixed-source result at step~50 is not evidence of a usable robust model. Benign assistance improves at steps~250 and~500, but those anchors also partially reopen under fixed-source and sparse adaptive attacks. We therefore use this audit as a check on the frontier interpretation, not as a broad capability benchmark.

\begin{table}[!htbp]
\centering
\scriptsize
\setlength{\tabcolsep}{2.5pt}
\renewcommand{\arraystretch}{0.94}
\begin{tabular}{@{}llccccc@{}}
\hline
Regime & Step & Strict & Lenient & Help. & Refusal & Empty \\
\hline
R2D2 & Reference & 0.000 & 0.000 & 0.000 & 1.000 & 0.000 \\
R2D2 & Step 50 & 0.000 & 0.000 & 0.000 & 1.000 & 0.000 \\
R2D2 & Step 250 & 0.150 & 0.633 & 0.783 & 0.367 & 0.000 \\
R2D2 & Step 500 & 0.250 & 0.867 & 1.117 & 0.133 & 0.000 \\
SFT & Reference & 0.050 & 0.750 & 0.800 & 0.200 & 0.100 \\
SFT & Step 50 & 0.233 & 0.933 & 1.167 & 0.017 & 0.050 \\
SFT & Step 250 & 0.283 & 0.983 & 1.267 & 0.000 & 0.000 \\
SFT & Step 500 & 0.217 & 1.000 & 1.217 & 0.000 & 0.000 \\
\hline
\end{tabular}
\caption{Direct benign-utility annotations on the 60-prompt sanity set. Strict utility counts fully helpful responses, lenient utility counts partially or fully helpful responses, and mean helpfulness averages the 0--2 annotation score. Refusal and empty-or-nonsensical outputs are reported as rates. This set is a sanity audit, not a broad capability benchmark.}
\label{tab:utility_baseline}
\end{table}

\subsection{Principal-Angle Drift Diagnostic}
\label{app:drift_diagnostic}

\begin{figure}[!htbp]
  \centering
  \includegraphics[width=0.94\linewidth,trim=6pt 6pt 6pt 6pt,clip]{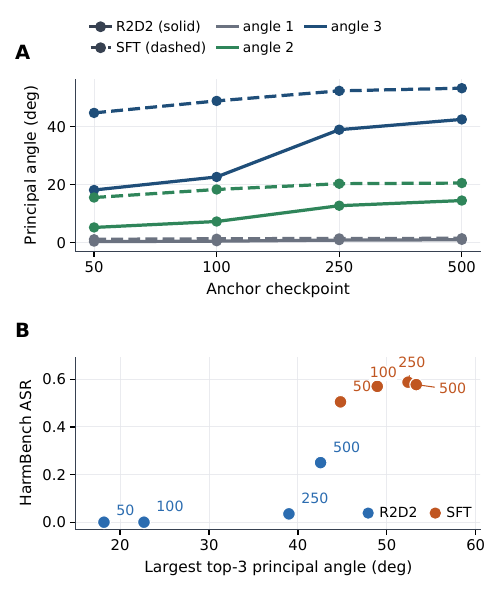}
  \caption{Principal-angle drift diagnostic. Panel A plots the top three principal angles between each anchor checkpoint and its regime-specific reference subspace; solid lines show R2D2 and dashed lines show SFT\@. Panel B compares the largest of these angles with fixed-source HarmBench GCG ASR, where lower ASR indicates stronger fixed-source robustness. SFT can drift more than R2D2 while remaining less robust, so drift magnitude alone is not sufficient to explain robustness.}
  \label{fig:drift_vs_robustness}
\end{figure}

Figure~\ref{fig:drift_vs_robustness} reports the full principal-angle diagnostic referenced in Section~\ref{sec:results-geometry}. This diagnostic provides a counterexample to a monotone ``larger rotation means stronger robustness'' account. SFT shows larger late-anchor rotation than R2D2, but remains substantially less robust under fixed-source HarmBench. We therefore treat principal-angle drift as a trajectory descriptor. The mechanism-level interpretation relies instead on which carrier is KL-admissible, where it is located, and how it behaves under intervention.

\subsection{Causal Intervention Side Effects}
\label{app:causal_side_effects}

\begin{figure}[!htbp]
  \centering
  \includegraphics[width=0.94\linewidth,trim=6pt 6pt 6pt 6pt,clip]{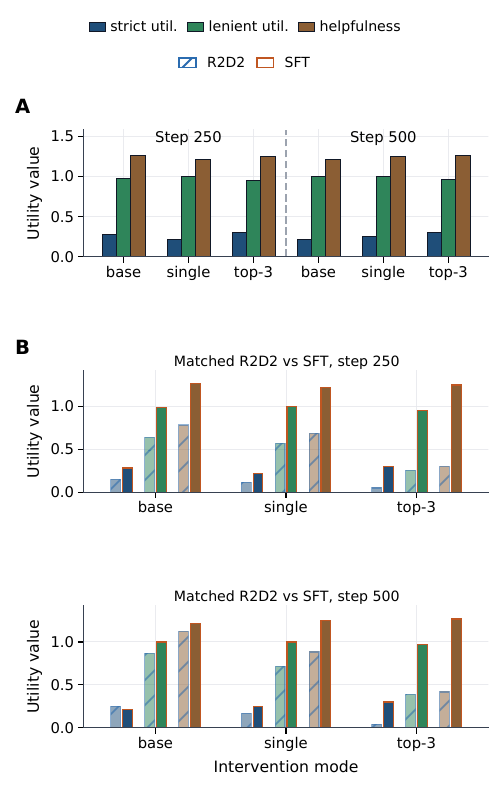}
  \caption{Utility effects of SFT and matched R2D2 interventions. Panel A shows SFT-only direct benign-utility metrics for the baseline, single-direction ablation, and top-3 subspace ablation at steps~250 and~500. Panel B overlays R2D2 and SFT utility metrics for the same intervention modes at each step; hatched bars denote R2D2 and outlined bars denote SFT\@. The contrast shows that the large late-stage utility cost is concentrated in the R2D2 intervention rows.}
  \label{fig:sft_intervention_utility}
\end{figure}

The intervention appendix reports the row-level behavior and utility costs behind the causal summary in Section~\ref{sec:results-causal}. Table~\ref{tab:causal_behavior} shows that late-stage R2D2 single-direction and top-3 ablations nearly eliminate fixed-source ASR, but also push XSTest refusal back upward. Table~\ref{tab:r2d2_intervention_utility} shows the matching benign-utility cost: stronger R2D2 interventions reduce strict and lenient utility while increasing benign refusal. Figure~\ref{fig:sft_intervention_utility} reports the matched SFT utility view and shows that SFT interventions do not incur the same late-stage utility collapse. Finally, Table~\ref{tab:step50_top3_sanity} resolves the step-50 ambiguity: top-3 ablation removes XSTest refusal without reopening fixed-source ASR, but the outputs become empty or nonsensical rather than useful. Taken together, these rows support low-dimensional but utility-coupled control, not a deployable intervention or fail-closed redundancy.

\begin{table}[!htbp]
\centering
\small
\renewcommand{\arraystretch}{0.94}
\resizebox{\columnwidth}{!}{%
\begin{tabular}{llccc}
\hline
Regime & Step & Mode & HarmBench ASR & XSTest any-refusal \\
\hline
R2D2 & 250 & Baseline & 0.0350 & 0.664 \\
R2D2 & 250 & Single direction & 0.0031 & 0.764 \\
R2D2 & 250 & Top-3 subspace & 0.0000 & 0.988 \\
R2D2 & 500 & Baseline & 0.2500 & 0.228 \\
R2D2 & 500 & Single direction & 0.0031 & 0.364 \\
R2D2 & 500 & Top-3 subspace & 0.0000 & 0.968 \\
SFT & 250 & Baseline & 0.5875 & 0.076 \\
SFT & 250 & Single direction & 0.5656 & 0.060 \\
SFT & 250 & Top-3 subspace & 0.4531 & 0.080 \\
SFT & 500 & Baseline & 0.5775 & 0.076 \\
SFT & 500 & Single direction & 0.5656 & 0.072 \\
SFT & 500 & Top-3 subspace & 0.4875 & 0.100 \\
\hline
\end{tabular}}
\caption{Behavior-side effects of causal ablations. Rows compare no-intervention baselines with single-direction and top-3 subspace ablations at matched R2D2 and SFT checkpoints. HarmBench ASR is fixed-source GCG attack success; XSTest any-refusal is the safe-prompt over-refusal rate. Late R2D2 ablations sharply reduce ASR but raise XSTest refusal.}
\label{tab:causal_behavior}
\end{table}

\begin{table}[!htbp]
\centering
\scriptsize
\setlength{\tabcolsep}{2.8pt}
\renewcommand{\arraystretch}{0.96}
\begin{tabular}{@{}llccccc@{}}
\hline
Step & Mode & \shortstack{Strict\\util.} & \shortstack{Lenient\\util.} & \shortstack{Mean\\help.} & \shortstack{Refusal\\rate} & \shortstack{Empty\\rate} \\
\hline
250 & Baseline & 0.150 & 0.633 & 0.783 & 0.367 & 0.000 \\
250 & Single dir. & 0.117 & 0.567 & 0.683 & 0.433 & 0.000 \\
250 & Top-3 sub. & 0.050 & 0.250 & 0.300 & 0.750 & 0.000 \\
500 & Baseline & 0.250 & 0.867 & 1.117 & 0.133 & 0.000 \\
500 & Single dir. & 0.167 & 0.717 & 0.883 & 0.267 & 0.017 \\
500 & Top-3 sub. & 0.033 & 0.383 & 0.417 & 0.617 & 0.000 \\
\hline
\end{tabular}
\caption{Benign-utility cost of late-stage R2D2 causal ablations on the 60-prompt benign set. Read with Table~\ref{tab:causal_behavior}, these rows show that stronger ablations trade lower fixed-source ASR for lower benign utility and higher benign refusal.}
\label{tab:r2d2_intervention_utility}
\end{table}

\begin{table}[!htbp]
\centering
\scriptsize
\setlength{\tabcolsep}{2.5pt}
\renewcommand{\arraystretch}{0.94}
\begin{tabular}{lccccc}
\hline
Mode & HB ASR & XS any-ref. & Strict & Lenient & Empty \\
\hline
Baseline & 0.000 & 1.000 & 0.000 & 0.000 & 0.000 \\
Top-3 subspace & 0.000 & 0.000 & 0.000 & 0.000 & 1.000 \\
\hline
\end{tabular}
\caption{R2D2 step-50 top-3 sanity check. Top-3 ablation removes XSTest refusal without reopening fixed-source ASR, but all utility scores remain zero and empty-or-nonsensical outputs reach 1.000.}
\label{tab:step50_top3_sanity}
\end{table}

\end{document}